\documentclass[journal]{IEEEtran}

\usepackage{amssymb}
\usepackage{amsthm}
\usepackage{bm} 
\usepackage{soul, color} 
\soulregister\cite7
\soulregister\ref7
\soulregister\pageref7

\usepackage{multirow} 
\usepackage{float} 
\usepackage{multicol}
\usepackage{siunitx} 
\usepackage{soul}
\usepackage[switch]{lineno} 

\usepackage{cite}

\usepackage[pdftex]{graphicx}
\graphicspath{{figures/}}
\DeclareGraphicsExtensions{.pdf,.jpeg,.png}

\usepackage[cmex10]{amsmath}
\usepackage{algorithmic}

\usepackage{array}

\ifCLASSOPTIONcompsoc
  \usepackage[caption=false,font=normalsize,labelfont=sf,textfont=sf]{subfig}
\else
  \usepackage[caption=false,font=footnotesize]{subfig}
\fi

\usepackage{url}
\PassOptionsToPackage{hyphens}{url}\usepackage{hyperref}

\begin{document}

\title{Hybrid Aerodynamics-Based Model Predictive Control for a Tail-Sitter UAV}

\author{ Bailun~Jiang,
         Boyang~Li,~\IEEEmembership{Member,~IEEE,}
         Ching-Wei~Chang, 
         and~Chih-Yung~Wen
         
\thanks{B. Jiang and C.-Y. Wen are with the AIRo-Lab, Department of Aeronautical and Aviation Engineering, The Hong Kong Polytechnic University, Hung Hom, Kowloon, Hong Kong. E-mail: (bailun-robin.jiang@connect.polyu.hk; bo-yang.li@polyu.edu.hk; cywen@polyu.edu.hk).}
\thanks{B. Li is with the School of Engineering, The University of Newcastle, Callaghan, NSW, Australia. E-mail: (boyang.li@newcastle.edu.au)}
\thanks{C.-W. Chang is with the Department of Mechanical Engineering, The Hong Kong Polytechnic University, Hung Hom, Kowloon, Hong Kong. E-mail: (chingwei.chang@connect.polyu.hk).}}

\maketitle

\begin{abstract}
It is challenging to model and control a tail-sitter unmanned aerial vehicle (UAV) because its blended wing body generates complicated nonlinear aerodynamic effects, such as wing lift, fuselage drag, and propeller--wing interactions. We therefore devised a hybrid aerodynamic modeling method and model predictive control (MPC) design for a quadrotor tail-sitter UAV. The hybrid model consists of the Newton--Euler equation, which describes quadrotor dynamics, and a feedforward neural network, which learns residual aerodynamic effects. This hybrid model exhibits high predictive accuracy at a low computational cost and was used to implement hybrid MPC, which optimizes the throttle, pitch angle, and roll angle for position tracking. The controller performance was validated in real-world experiments, which obtained a 57\% tracking error reduction compared with conventional nonlinear MPC. External wind disturbance was also introduced and the experimental results confirmed the robustness of the controller to these conditions.
\end{abstract}


\section*{Supplementary Materials}
The collected dataset for model training can be found at \url{https://ieee-dataport.org/documents/quadrotor-tail-sitter-uav-flight-log}. A video demonstration is available at \url{https://youtu.be/GLJdBFR3OvU}. The implemented algorithm source code is availiable at \url{https://github.com/HKPolyU-UAV/airo_control_interface}.

\section{INTRODUCTION}
R{\scshape ecent} advancements in microcontroller and sensing technologies have led to rapid developments in unmanned aerial vehicles (UAVs). The UAVs find extensive application in tasks such as environmental monitoring, land surveying, and disaster response \cite{nikolic2013uav,siebert2014mobile, sun2016camera}. These applications impose specific requirements on the performance of UAV systems, including factors such as endurance during flight, tracking accuracy, and wind resistance capability.

\begin{figure}[thpb]
    \centering
      \includegraphics[width=0.48\textwidth]{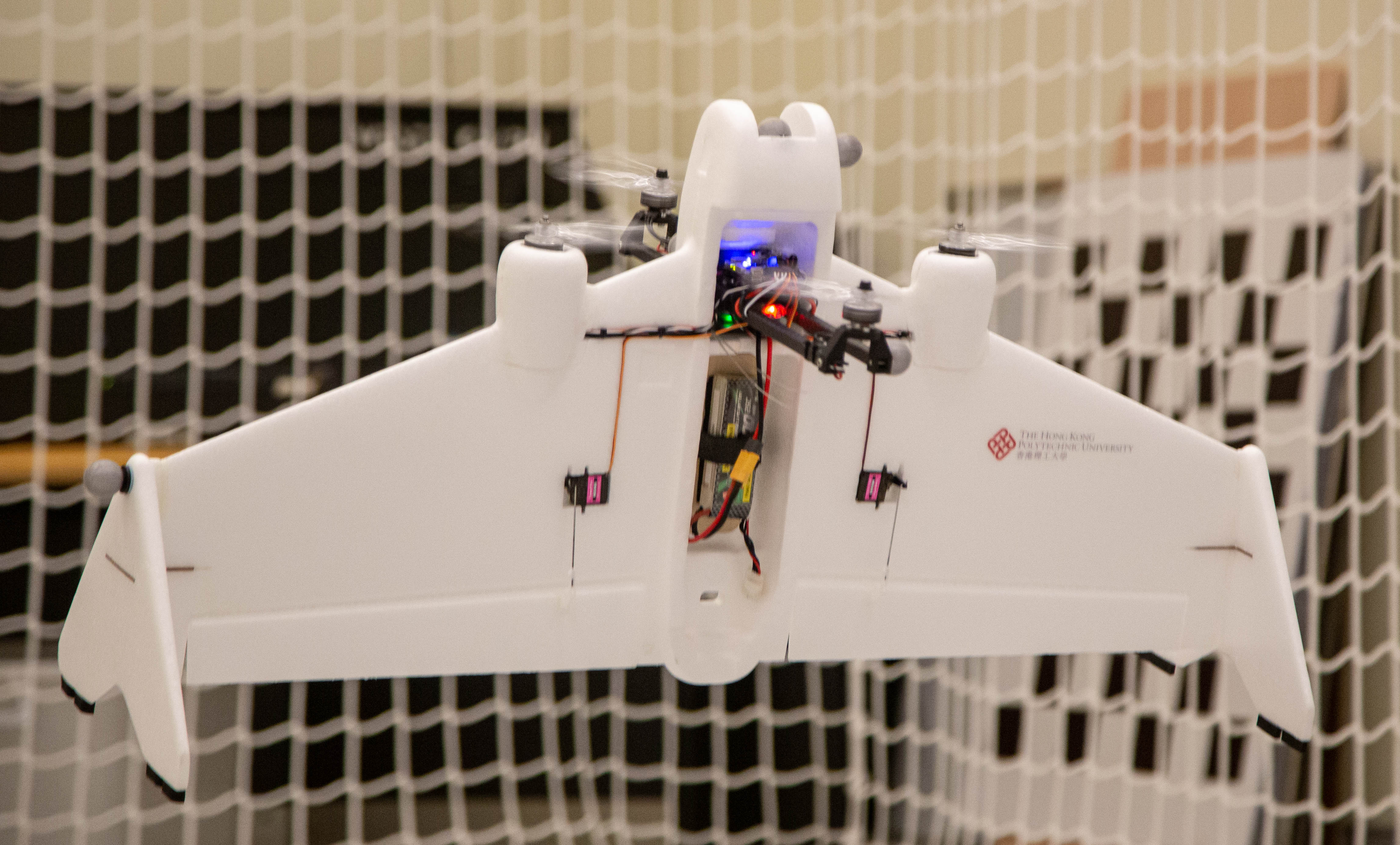}
      \caption{The tail-sitter quadrotor UAV in hovering flight against a strong wind disturbance.}
      \label{arkbird}
\end{figure}

Multi-rotor UAVs have simple structures and agile maneuvering capabilities. However, their flight range and endurance are usually limited by their relatively low aerodynamic efficiency. Conversely, fixed-wing UAVs characterized by high eficiency typically necesitate a runway for both take-off and landing, a condition that may not be feasible in urban environments. Therefore, versatile hybrid UAVs have been developed that can convert between rotary-wing and fixed-wing modes to balance the trade-off between flight efficiency and vertical take-off and landing (VTOL) capabilities, such as tail-sitter, quad-plane, and tilt-wing UAVs \cite{bauersfeld2021mpc, ke2018design, li2020transition, benkhoud2016model, li2021dynamic, bapst2015design}. Such VTOL UAVs usually take-off and land in urban area with limited space or on moving platforms such as boats. In these conditions, precise position tracking and robustness against disturbance are crucial. For transition and level-flight modes, on the other hand, the tracking accuracy is not as important as the hovering phase since these maneuvers are usually conducted up in the air with few obstacles.

This study focused on the hovering control of a quadrotor tail-sitter UAV based on the dualrotor Arkbird VTOL airframe, shown in Fig. \ref{arkbird}. The quadrotor tail-sitter UAV is more controllable and robust than its dualrotor counterpart. The large wing area of the quadrotor tail-sitter UAV mean that its dynamics are dominated by the strong nonlinear aerodynamic effects of the wing, which increases the difficulty of flight controller design. In hovering flight, the quadrotor tail-sitter UAV shares a similar control form to that of a quadrotor UAV; herein, its pitch and yaw control inputs were mapped to elevon angle command to enhance its controllability.

Accurate trajectory tracking plays a pivotal role in enhancing the operational safety and efficiency of UAVs across diverse environmental conditions. Many research works have developed and implemented model predictive control (MPC) to enhance the position tracking accuracy on UAVs\cite{bouffard2012board, bangura2014real, Kamel2017}. MPC considers a system dynamic model based on the receding horizon principle. Consequently, it leverages the model to anticipate the future dynamics of a control process and calculate optimal control inputs while adhering to state constraints. The first value of the optimal control input sequence derived by the optimizer will be applied on the system. This iterative online optimization process is executed at each time-step, allowing for continuous refinement of control inputs based on real-time information and system dynamics. MPC offers an additional benefit in comparison to traditional reactive controllers like the proportional-integral-derivative (PID) controller. Unlike the latter, which corrects states solely based on tracking error and consequently leads to tracking delays, MPC employs proactive control actions on the system. This proactive approach enables MPC to respond to future set-points and disturbances within the prediction horizon, enhancing its adaptability and responsiveness.

Model-based control methods such as MPC can be substantially enhanced by the use of an accurate system model. However, it is almost impossible to derive a perfect mathematical model that comprehensively describes the complete system dynamics of UAVs in real-world settings. MPC algorithms that are currently applied to tail-sitter UAVs typically employ a basic quadrotor model linearized around hovering conditions and neglect residual aerodynamic effects such as wing lift and fuselage drag \cite{li2018development,zhou2019position}. As the velocity of a UAV increases, this neglect of residual aerodynamics has an increasing effect on the UAV and can lead to significant tracking errors.

A neural network (NN) can be used to model these residual aerodynamic effects, as NNs are universal function approximators that are capable of modeling highly nonlinear dynamic systems. Therefore, various NN-based methods have been developed for the modeling and control of UAVs  \cite{mohajerin2018deep, mohajerin2015modelling, bansal2016learning}. Given that the neural network serves as the prediction model, the complexity of the model is constrained by the control frequency and the computational capability of the platform. Deep recurrent neural networks (RNNs) exhibit the capability to learn explicit system behaviors, including features like motor delay and airflow states. However, deploying such RNNs in real-time for optimal control calculations poses challenges to the system. Consequently, a crucial consideration in the modeling process involves striking a balance between model accuracy and simplicity to address the trade-off inherent in this context.
 
In the current study, hybrid MPC (HMPC) was devised for position tracking of a quadrotor tail-sitter UAV in hovering flight. The flight data of the UAV were collected by commanding it to follow trajectories that excited translational and rotational movements in all axes. A feedforward neural network (FFNN) was trained to learn the residual aerodynamic effects and was used as a compensator for the nonlinear quadrotor model. The model was implemented with HMPC to predict the states of the UAV. We compared the performance of HMPC with that of nonlinear MPC (NMPC) using the nonlinear quadrotor model with trajectories not included in the training data. We also introduced wind disturbances to test the robustness of the controller. The main contributions of this study are as follows:

\begin{itemize}
	{\item an FFNN with good accuracy and a simple structure was used to learn the residual aerodynamic effects of a quadrotor tail-sitter UAV from its flight data;
	\item by combining the learned dynamics with Newton-Euler model, a position controller was developed within an MPC pipeline and allows for update frequencies greater than 100Hz; and
	\item HMPC was demonstrated to be robust and outperformed NMPC by up to 57\% in real-world flight experiments.}
\end{itemize}

The remainder of this paper is organized as follows. Section II presents a review of related studies. Section III introduces the governing equations of quadrotor tail-sitter UAV and describes the development of an NN modeling method. Section IV details the testing of controller performance in real-world experiments Section V concludes. 

\section{Related Works}

Versatile tail-sitter UAV configurations have been designed and their dynamic models have been established via various approaches. Based on simplified low-speed aerodynamic principles, Knoebel et al. \cite{knoebel2006preliminary} modeled a tail-sitter UAV in hovering flight by considering its propellers and control surfaces separately. Stone \cite{stone2008aerodynamic} used a full azimuthal base-element solution to predict the aerodynamic forces for the propellers of a tail-sitter UAV and combined the predictions with a panel method model. Sun et al. \cite{sun2017dynamic} introduced a hardware-in-the-loop (HITL) simulation platform for a dual-rotor tail-sitter UAV, relying on an aerodynamic model derived from wind tunnel experiments covering the full flight envelope for this UAV. Lyu et al. \cite{lyu2018simulation} performed a full attitude wind-tunnel test on a full-scale UAV to accurately capture the degradation of motor thrust and torque in the presence of forward speed. 

An MPC scheme can be established by synthesizing the determined dynamics into a prediction model. Li et al. \cite{li2018development} developed linear MPC for a tail-sitter UAV in hovering flight. Although the tail-sitter aerodynamics and actuator were modeled in detail, the MPC afforded a linearized quadrotor dynamics model. Following this work, Zhou et al. \cite{zhou2019position} devised a successive linearization method for a tail-sitter UAV MPC scheme, with the prediction model linearized around the UAV's current attitude to attenuate prediction errors. These approaches have been validated in relatively low-speed scenarios (approximately 0.5 \si{m/s}), in which quadrotor dynamics have a nominal effect on a system. However, these linear models fail to predict the dynamics for higher-speed scenarios, which can lead to large tracking errors. This underscores the need for accurate MPC for tail-sitter UAVs, especially for higher-speed scenarios.

NNs are capable of precisely modeling aerodynamic effects with excellent computational efficiency, and many recent studies have employed NNs to learn quadrotor dynamics models from data. Bansal et al. \cite{bansal2016learning} modeled quadrotor dynamics using an FFNN known as a rectified linear unit, which they employed in a linear–quadratic regulator controller. Torrente et al. \cite{torrente2021data} modeled aerodynamic effects with Gaussian processes and incorporated these with MPC to achieve efficient and precise real-time position control. In real-world experiments, they achieved flight speeds of up to 14 \si{m/s} and accelerations beyond 4\si{g}. Bauersfeld et al. \cite{bauersfeld2021neurobem} modeled quadrotor aerodynamics with blade-element theory and learned residual dynamics with NNs. The resulting model captures aerodynamic thrust, torques, and parasitic effects at speeds up to 18 \si{m/s} and showed strong generalization capabilities beyond the training set. As a general quadrotor UAV has a relatively small aerodynamic area, nonlinear aerodynamic effects only become dominant at high speeds (usually $>$ 5 \si{m/s}). However, for a tail-sitter UAV, such effects are non-negligible even at low speeds.

Inspired by \cite{bansal2016learning} and \cite{bauersfeld2021neurobem}, we developed a method that combines a nonlinear quadrotor model based on a Newton--Euler formalism with an NN-based method to learn residual aerodynamic effects. This method greatly simplifies the process of modeling of a tail-sitter UAV, without the need for wind tunnel tests. The resulting HMPC inherits the generalization capability of the Newton--Euler model and the function-approximation feature of an NN.

\section{Methodology}

This section introduces the dynamic equations, the NN model, and the MPC implementation for the quadrotor tail-sitter UAV. We fully exploited the inherently proactive features of MPC by concentrating on outer loop control, as only a trajectory preview was available for the position of the UAV.

\subsection{Dynamic Model of the Quadrotor Tail-Sitter UAV\label{dynamics_model}}

\begin{figure}[thpb]
    \centering
      \includegraphics[width=0.3\textwidth]{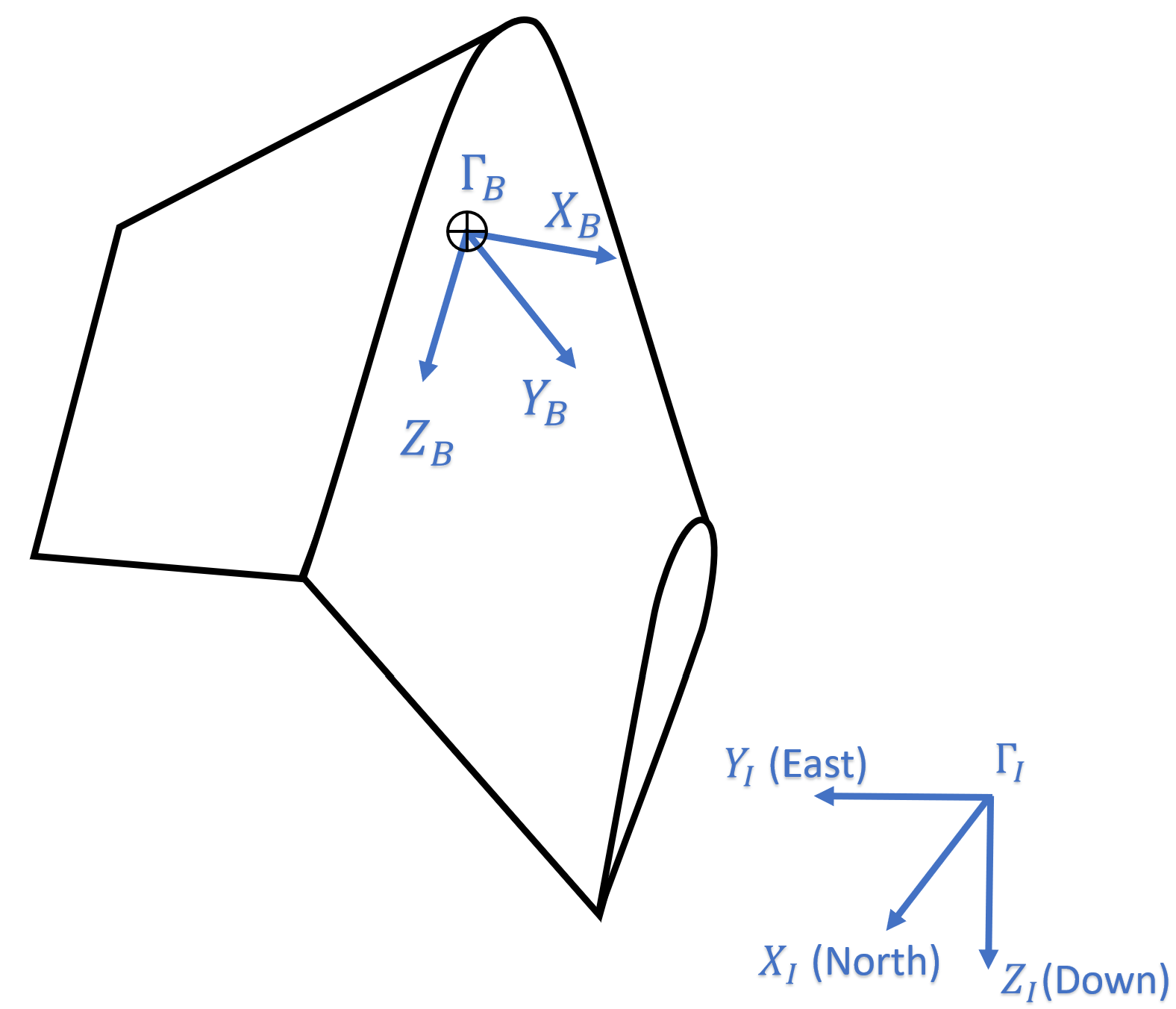}
      \caption{Schematic illustration of the quadrotor tail-sitter UAV inertial frame $\Gamma_I$ and body frame $\Gamma_B$.}
      \label{coordinate_frame}
\end{figure}

To model the dynamics of the tail-sitter UAV, we assume it to be a 6-DOF rigid body which is defined in both inertial frame $\Gamma_I$ and body frame $\Gamma_B$, as shown in Fig. \ref{coordinate_frame}. The tail-sitter UAV position, velocity, and acceleration in $\Gamma_I$ are defined by $\mathbf{p} = \left[x\ y\ z\right]^\text{T}\in \mathbb{R}^{3}$, $\mathbf{v} = \left[u\ v\ w\right]^\text{T}\in \mathbb{R}^{3}$, and $\mathbf{a} = \left[a_x\ a_y\ a_z\right]^\text{T}\in \mathbb{R}^{3}$. Its orientation is denoted by $\mathbf{R}_\text{B}^\text{I}\in \text{SO}(3)$ and its angular velocity is denoted by $\boldsymbol{\omega}$.

The kinetic and dynamic model of the tail-sitter UAV is constructed based on Newton--Euler formalism, as follows
\begin{linenomath}\begin{equation}
	\label{netwon-euler}
	\begin{aligned}
		\dot{\mathbf{p}}&=\mathbf{v}\\
		m\dot{\mathbf{v}}&=\mathbf{R}_\text{B}^\text{I}\mathbf{F}_\text{B}\\            \dot{\mathbf{R}}_\text{B}^\text{I}&=\mathbf{R}_\text{B}^\text{I}\boldsymbol{\omega}_{\times}\\
            \mathbf{I}\dot{\boldsymbol{\omega}}&=-\boldsymbol{\omega}\times(\mathbf{I}\boldsymbol{\omega})+\mathbf{M}_\text{B},\\
	\end{aligned}
\end{equation}\end{linenomath}
where $m$ is the mass and $\mathbf{I}$ is the inertial matrix of the tail-sitter UAV. $\boldsymbol{\omega}_{\times}$ is the skew-symmetric matrix, such that $\boldsymbol{\omega}_{\times}\mathbf{V}=\boldsymbol{\omega}\times\mathbf{V}$ for any vector $\mathbf{V}\in \mathbb{R}^3$. The total force and moments applied in $\Gamma_B$ are denoted by $\mathbf{F}_\text{B}$ and $\mathbf{M}_\text{B}$, which can be expanded as follows:
\begin{linenomath}\begin{equation}
    \label{force_moment}
    \begin{aligned}        \mathbf{F}_\text{B}&=\mathbf{F}_\text{prop}+\mathbf{F}_\text{aero}+\mathbf{R}_\text{I}^\text{B}\mathbf{F}_\text{g}+\mathbf{F}_\text{d}\\        \mathbf{M}_\text{B}&=\mathbf{M}_\text{prop}+\mathbf{M}_\text{aero}+\mathbf{M}_\text{d},\\
    \end{aligned}
\end{equation}\end{linenomath}
where $\mathbf{F}_\text{prop}$ and $\mathbf{M}_\text{prop}$ denote the force and moment generated by the propellers, respectively; $\mathbf{F}_\text{aero}$ and $\mathbf{M}_\text{aero}$ denote the force and moments generated by the aerodynamic effects, respectively; $\mathbf{F}_\text{d}$ and $\mathbf{M}_\text{d}$ denote the external disturbance force and moments, respectively; and $\mathbf{F}_\text{g}$ is the gravitational force vector. 

\subsection{NN--Augmented Hybrid Model}

$\mathbf{F}_\text{aero}$ and $\mathbf{M}_\text{aero}$ may be the dominant terms within the overall dynamics of a tail-sitter UAV, especially during high-speed maneuvers. In addition to wing lift and drag, other effects such as fuselage drag, propeller--wing interactions, and rotor-to-rotor interactions also contribute to $\mathbf{F}_\text{aero}$ and $\mathbf{M}_\text{aero}$ \cite{knoebel2006preliminary} \cite{stone2008aerodynamic} . The tracking performance of MPC can be substantially improved by considering these aerodynamic effects. However, these effects are either too complicated to be described by mathematical equations or necessitate extensive experiments (such as computational fluid dynamics (CFD) or wind tunnel tests) to identify the coefficients for their mathematical equations. In a previous study \cite{li2018development}, we developed a wing lift and drag force model for a tail-sitter UAV based on wind tunnel experiments. This model can be employed in the HITL simulations but not in MPC. In the current study, we used an NN approach to model the aerodynamic effects of a tail-sitter UAV purely from its flight data. The model shows good accuracy yet is lightweight enough for real-time calculation of optimal control problems.

In the hybrid aerodynamic model, the basic dynamics generated by quadrotor propellers and gravity are described by a nonlinear model based on a Newton--Euler formalism. The residual aerodynamic effects and the potential error of the quadrotor model are approximated by an NN. As we focused on the outer loop position control, we modeled only translational dynamics. Therefore, the training target was to minimize the difference between the actual acceleration and the acceleration predicted by the nonlinear model. The network inputs are linear velocity, attitudes, and collective thrust. Position and angular velocity are excluded as inputs, given that they are not directly associated with translational dynamics. The NN model features 6 inputs and 3 outputs, expressed as follows:
\begin{linenomath}\begin{equation}\label{networkfunction}
	\hat{\mathbf{a}}_\text{res} = f_\text{NN}(u,v,w,\phi,\theta,T),
\end{equation}\end{linenomath}
where $\hat{\mathbf{a}}_\text{res} = [\hat{a}_{x,\text{res}}\ \hat{a}_{y,\text{res}}\ \hat{a}_{z,\text{res}}]^\text{T}$ represents the predicted residual accelerations and $f_\text{NN}$ represents the NN transfer function.

To predict the residual dynamics of the tail-sitter, a multilayer perceptron (MLP) known as FFNN with similar structure as the ReLU network model in \cite{punjani2015deep,bansal2016learning} is used. Several different NN model has been used to predict the UAV dynamics. RNN used in \cite{punjani2015deep, mohajerin2019multistep, mohajerin2015modelling} allows the NN to learn the dynamics of a more complex system based on the past state information. However, this makes it hard to train the network and design the controller \cite{bansal2016learning}. Therefore, a feed-forward fashion is chosen to ensure the online computational capability within an MPC pipeline. The performances of two types of feed-forward NNs temporal-convolutional (TCN) encoders and MLP are compared in \cite{bauersfeld2021neurobem}. The result shows that with similar sizes, MLP has better torque prediction and similar force prediction as TCN. Due to its simple structure and relatively accurate prediction result, FFNN is used in this work.

The two-layer FFNN model comprises one hidden layer and one output layer. It utilizes sigmoid and linear transfer functions for the hidden and output layers, respectively. The network model can be represented as follows:
\begin{linenomath}\begin{equation}\label{networkmodel}
	\boldsymbol{\alpha} = \boldsymbol{w}^\text{T}\xi\left(\boldsymbol{W}^\text{T}\boldsymbol{\beta}+\boldsymbol{B}\right)+\boldsymbol{b},
\end{equation}\end{linenomath}
where $\boldsymbol{\alpha}$ and $\boldsymbol{\beta}$ represent the network output and input vector, respectively. The weight matrices are represented as $\boldsymbol{W}$ and $\boldsymbol{w}$, with the corresponding bias vectors denoted by $\boldsymbol{B}$ and $\boldsymbol{b}$. Here, uppercase letters signify parameters related to the hidden layer, while the lowercase letters represent those associated with the output layer. Additionally, $\xi$ denotes the sigmoid activation function applied in the hidden layer, while the output layer adopts the linear activation function.

The training process follows a supervised learning approach, employing the Levenberg--Marquardt algorithm to optimize the weights and biases. The objective is to minimize the mean squared error (MSE) between the target and the output expressed as follows:
\begin{linenomath}\begin{equation}\label{mse}
	\min\sum_{k=1}^{N}\frac{1}{N}\left \| \hat{\mathbf{a}}_\text{res}-\mathbf{a}_\text{res} \right \| ^ 2,
\end{equation}\end{linenomath}
where $\mathbf{a}_\text{res} = \mathbf{a} - \mathbf{a}_\text{nonlin}$ denotes the target residual dynamics, which equal the difference between actual acceleration and that predicted by the nonlinear quadrotor model $\mathbf{a}_\text{nonlin}$; and $N$ denotes the number of training data.

\subsection{HMPC Implementation\label{HMPC}}

\begin{figure}[thpb]
\centering
\includegraphics[width=0.48\textwidth]{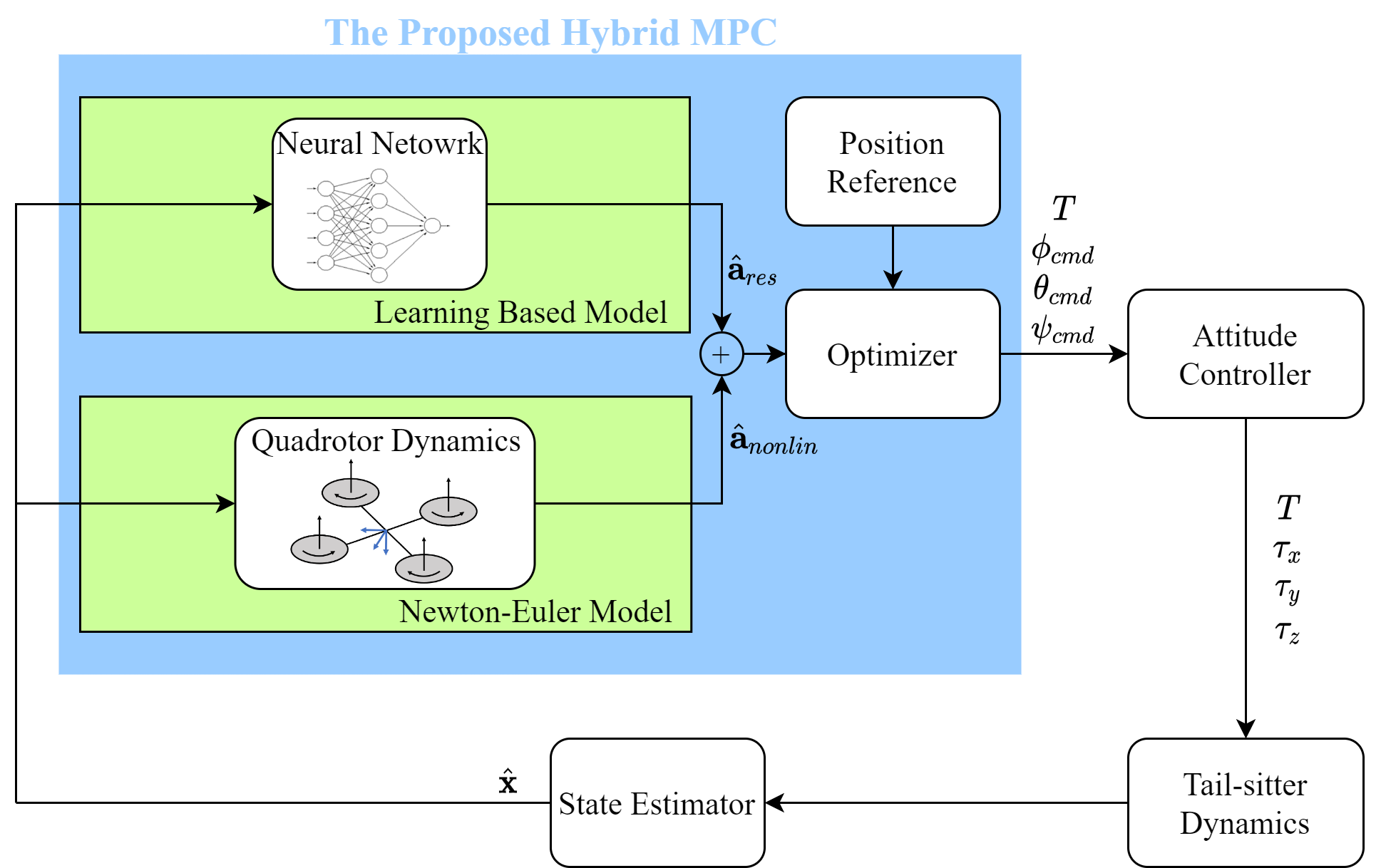}
\caption{Diagram of the cascaded control architecture, where the green blocks indicate the Newton--Euler and learning based models and blue block indicates the proposed HMPC architecture.}
\label{hmpc_architecture}
\end{figure}

The trained FFNN mentioned in previous section is integrated into a hybrid model that combines the nonlinear and neural network models, which serves as prediction model for the HMPC. The proposed HMPC is implemented in cascaded loops, with the outer loop designed for positional control and the inner loop designed for attitudinal control. In position tracking tasks, the future reference trajectory is usually available several seconds in advance, whereas the reference attitude is calculated in real-time. This indicates that the preview feature is only applicable to outer-loop positional control. To fully exploit the preview feature of MPC, we focused on positional control of the quadrotor tail-sitter UAV. The architecture of the hybrid aerodynamic model and cascaded control loop is shown in Fig. \ref{hmpc_architecture}, where $\hat{\mathbf{x}}$ denotes the estimated UAV states. There are two components in the HMPC model: a nonlinear quadrotor dynamics model based on a Newton--Euler formalism and a learning-based model describing residual dynamics. The predicted result is the sum of outputs from these two models. The optimizer calculates the attitude and collective thrust commands that minimize the difference between the predicted and reference position, which are then sent to the inner loop for attitude tracking.

The state of HMPC is defined by $\mathbf{x}=\left [ x\ y\ z\ u\ v\ w\ \phi\ \theta\ \right ]^\text{T}$, where $\phi$ and $\theta$ denote the roll and pitch angles in the body frame $\Gamma_\text{B}$, respectively. The control input is defined by $\mathbf{u}=[T\ \phi_\text{cmd}\ \theta_\text{cmd}]^\text{T}$, where $T$ is the collective thrust force generated by the actuators, and $\phi_\text{cmd}$ and $\theta_\text{cmd}$ are roll and pitch commands, respectively, which are sent to the inner loop controller. The hybrid prediction model is formulated by
\begin{linenomath}\begin{equation}
	\label{prediction_model}
	\left\{
	\begin{aligned}
		\dot x&=u\\
		\dot y&=v\\
		\dot z&=w\\
		\dot u&=\cos\phi \sin\theta \ \frac{T}{m}+\hat{a}_{x,\text{res}}\\
		\dot v&=-\sin\phi \ \frac{T}{m}+\hat{a}_{y,\text{res}}\\
		\dot w&=-g+\cos\phi \cos\theta \ \frac{T}{m}+\hat{a}_{z,\text{res}}\\
		\dot{\phi}&=\frac{\phi_\text{cmd}-\phi}{\tau_\phi}\\
		\dot{\theta}&=\frac{\theta_\text{cmd}-\theta}{\tau_\theta}
	\end{aligned}
	\right.,
\end{equation}\end{linenomath}
where $g$ is the gravitational acceleration. The translational dynamics consist of nonlinear quadrotor dynamics and an NN model. A first-order model is used to approximate the inner loop dynamics, where $\tau_\phi$ and $\tau_\theta$ are time constants of roll and pitch control, respectively. The values of drag coefficients and controller time constants are identified from experimental data. The yaw command angle is set to zero throughout the experiment, such that the yawing movement is not considered in the dynamics model. 

The MPC optimizer addresses the quadratic programming problem, formulated as follows:
\begin{linenomath}\begin{equation}\label{qp}
	\begin{aligned}
		\min \quad &\int_{t=0}^{t_N} ||h(\mathbf{x}(t),\mathbf{u}(t))-\mathbf{y}_\text{ref} ||_\mathbf{Q}^2 dt\\
		&\quad \quad + || h(\mathbf{x}(t_{N}))-\mathbf{y}_{N,\text{ref}} ||_{\mathbf{Q}_N}^2 dt
		\\
		s.t.\quad &\dot{\mathbf{x}}=f(\mathbf{x}(t),\mathbf{u}(t))\\
		&\mathbf{u}(t)\in \mathcal{U}\\
		&\mathbf{x}(t)\in \mathcal{X}\\
		&\mathbf{x}(0)=\mathbf{x}(t_0),\\
	\end{aligned}
\end{equation}\end{linenomath}
where $\mathbf{u}(t)$ and $\mathbf{x}(t)$ denote control inputs and states at time-step $t$, respectively; $t_0$ and $t_N$ denote the first and last time-steps of the prediction horizon, respectively; $\mathbf{y}_\text{ref}$ and $\mathbf{y}_{N,\text{ref}}$ denote the reference state for the prediction horizon and the terminal time-step, including desired position, velocity, attitude and control inputs; $\mathbf{Q}$ and $\mathbf{Q}_N$ represent weighting matrices for states and terminal states, respectively; $f(\cdot)$ denotes the prediction function, and $h(\cdot)$ denotes the system output function; and $\mathcal{U}$ and $\mathcal{X}$ denote the input constraint and state constraint, respectively.

The OCP presented in (\ref{qp}) solved using multiple shooting method. A boundary value problem (BVP) is formulated with the discretized system and solved by employing the active set method in qpOASES solver\cite{Kamel2017,ferreau2014qpoases}. 

\section{Flight Experiments}

To validate the effectiveness of HMPC, we conducted real-world indoor flight experiments. The measurement data were recorded during several flights and then post-processed by time synchronization and noise filtering. 

\subsection{Performance Benchmarks}\label{performance_benchmarks}
To test the performance of proposed HMPC, multiple benchmark controllers are introduced, including PID,backstepping, sliding-mode, and NMPC. The backsteping and sliding-mode are involved to compare the performance of MPC against other commonly used nonlinear control methods, while the NMPC is involved to validate the effect of hybrid model compared to basic nonlinear model. The backstepping control law is described by\cite{bouabdallah2005backstepping}
\begin{linenomath}\begin{equation}
	\label{backstepping_control_law}
	\left\{
	\begin{aligned}
		T&=\frac{m}{\cos\phi\cos\psi}(e_z+\ddot z_d+k_{bz1}\dot e_z+g+k_{bz2}e_z)\\
		u_x&=\frac{m}{T}(e_x+\ddot x_d+k_{bx1}\dot e_x+g+k_{bx2}e_x)\\
		u_y&=\frac{m}{T}(e_y+\ddot y_d+k_{by1}\dot e_y+g+k_{by2}e_y)\\
	\end{aligned}
	\right.,
\end{equation}\end{linenomath}
with
\begin{linenomath}\begin{equation}
	\label{backstepping_variables}
	\left\{
	\begin{aligned}
		u_x&=\cos\phi_{cmd}\sin\theta_{cmd}\cos\psi + \sin\phi_{cmd}\sin\psi\\
		u_y&=\cos\phi_{cmd}\sin\theta_{cmd}\sin\psi - \sin\phi_{cmd}\cos\psi\\
            e_z&= z_d - z\\
            \dot e_z&=\dot z_d - \dot z\\
            e_x&= x_d - x\\
            \dot e_x&=\dot x_d - \dot x\\
            e_y&= y_d - y\\
            \dot e_y&=\dot y_d - \dot y\\
	\end{aligned}
	\right.,
\end{equation}\end{linenomath}
where $k_{b}$ is controller gain, $e$ represents tracking error in corresponding direction, $u_x$ and $u_y$ represent virtual control inputs for translational movements, subscript $d$ represents the desired position in corresponding direction.

The sliding-mode control law is described by\cite{bouabdallah2005backstepping}
\begin{linenomath}\begin{equation}
	\label{sliding_mode_control_law}
	\left\{
	\begin{aligned}
		T&=\frac{m}{\cos\phi\cos\psi}(k_{sz1}\dot e_z+\ddot z_d+g+k_{sz2}sgn(s_z))\\
		u_x&=\frac{m}{T}(k_{sx1}\dot e_x+\ddot x_d+k_{sx2}sgn(s_x))\\
		u_y&=\frac{m}{T}(k_{sy1}\dot e_y+\ddot y_d+k_{sy2}sgn(s_y))\\
	\end{aligned}
	\right.,
\end{equation}\end{linenomath}
with
\begin{linenomath}\begin{equation}
	\label{sliding_mode_variables}
	\left\{
	\begin{aligned}
		u_x&=\cos\phi_{cmd}\sin\theta_{cmd}\cos\psi + \sin\phi_{cmd}\sin\psi\\
		u_y&=\cos\phi_{cmd}\sin\theta_{cmd}\sin\psi - \sin\phi_{cmd}\cos\psi\\
            s_z&= k_{sz1}e_z+\dot e_z\\
            e_z&= z_d - z\\
            \dot e_z&=\dot z_d - \dot z\\
            s_x&= k_{sx1}e_x+\dot e_x\\
            e_x&= x_d - x\\
            \dot e_x&=\dot x_d - \dot x\\
            s_y&= k_{sy1}e_y+\dot e_y\\
            e_y&= y_d - y\\
            \dot e_y&=\dot y_d - \dot y\\
	\end{aligned}
	\right.,
\end{equation}\end{linenomath}
where $k_{s}$ is controller gain, $e$, $u_x$ and $u_y$ are defined by the same way as the backstepping controller, $s$ represents the sliding surfaces. The controller gains are summarized by Table \ref{backstepping_sliding_mode_parameters}.

\begin{table}[hbt!]
	\renewcommand{\arraystretch}{1}
	\caption{Backstepping and sliding-mode controller parameters}
	\label{backstepping_sliding_mode_parameters}
 \begin{center}
	\begin{tabular}{cc}
		\hline 
		Parameters & Value \\
		\hline 
		$[k_{bx1}$ $k_{bx2}]$ & [1 1]\\
            $[k_{by1}$ $k_{by2}]$ & [1 1]\\
            $[k_{bz1}$ $k_{bz2}]$ & [5 1]\\
		$[k_{sx1}$ $k_{sx2}]$ & [2 1]\\
            $[k_{sy1}$ $k_{sy2}]$ & [2 1]\\
            $[k_{sz1}$ $k_{sz2}]$ & [3 2.5]\\
		\hline 
	\end{tabular} 
  \end{center}
\end{table}

The NMPC benchmark predicts the system dynamics by a nonlinear Newton--Euler model given by

\begin{linenomath}\begin{equation}
	\label{prediction_model_nonlinear}
	\left\{
	\begin{aligned}
		\dot x&=u\\
		\dot y&=v\\
		\dot z&=w\\
		\dot u&=\cos\phi \sin\theta \ \frac{T}{m}+c_{dx}u|u|\\
		\dot v&=-\sin\phi \ \frac{T}{m}+c_{dy}v|v|\\
		\dot w&=-g+\cos\phi \cos\theta \ \frac{T}{m}+c_{dz}w|w|\\
		\dot{\phi}&=\frac{\phi_\text{cmd}-\phi}{\tau_\phi}\\
		\dot{\theta}&=\frac{\theta_\text{cmd}-\theta}{\tau_\theta}\\
	\end{aligned}
	\right..
\end{equation}\end{linenomath}

Note that compared to the hybrid model proposed in (\ref{prediction_model}), the nonlinear model uses additional quadratic terms to predict the translational drag forces, and the drag coefficients $c_{dx}$, $c_{dy}$, and $c_{dz}$ are derived with system identification technique.

The tracked trajectories are not included in the training dataset. In addition, we introduced wind disturbance to test the robustness of HMPC.

\subsection{Experimental Setup\label{experiment_setup}}

The real-world experiment was conducted in an indoor flight area equipped with a Vicon motion tracking system that provided data on the quadrotor tail-sitter UAV position and orientation at 300 \si{Hz}. The UAV was a dualrotor Arkbird VTOL airframe modified with two extra rotors fitted to beams across the fuselage, resulting in a total mass of 0.83 \si{kg}. The UAV was equipped with a 3S 2,200-mAh lithium-ion polymer battery, four Sunnysky R2207 KV2580 motors with electric speed controllers, four Racerstar T7056 three-bladed propellers, and two servo motors for elevon control. The quadrotor propeller diagonal distance was 33 \si{cm}. The autopilot hardware we used is PixRacer. It communicates with an onboard computer LattePanda Alpha, which has an Intel Core M3-8100Y CPU and an Ubuntu 20.04 OS. All the control algorithms were running on the onboard computer.

\subsection{Data Collection and Network Training}

We collected UAV flight data using both offboard and manual flight modes. In the offboard mode, the UAV tracked the local position references sent to MAVROS topic $~setpoint\_position/local$. The reference trajectory contained step commands in all three directions, and three step sizes were designed (1, 1.5, and 2 m). In manual mode, the UAV was operated by an expert pilot using Futaba T16SZ remote control. Three flight modes (position, altitude, and stabilized modes) were chosen to enrich the maneuvering intensity. Each experiment was performed three times to attenuate the influence of random error, resulting in 18 flight logs. The take-off and landing parts were excluded to improve the data quality. The total flight time was 24 min, and the position, orientation, and actuator control inputs were recorded onto the onboard storage card.

As the states of the UAV were recorded by various topics with different sampling rates, the raw data and timestamps were synchronized by interpolation at 100 Hz. All data were filtered using a moving average filter to remove high-frequency noise. The unobserved velocity and acceleration were calculated by differentiating the filtered data rather than the raw data, as differentiation amplifies discrete noise. The resulting dataset contained approximately 150,000 sample points for NN training.

The network uses translational velocities, Euler angles, and thrust as inputs, while the target outputs are residual translational accelerations. The dataset was partitioned, with 70\% used for training, 15\% for validation, and the remaining 15\% for testing. Each subset of data encapsulated the entire flight envelope present in the full dataset.

Training was performed with the MATLAB Deep Learning Toolbox \cite{matlab} using the Levenberg--Marquardt backpropagation algorithm to minimize the root-mean-square error (RMSE) of the predicted and target acceleration. Table \ref{train_rmse} summarizes the accelerations predicted by the nonlinear model and the hybrid models with various sizes of hidden layers, where the maximum solving times of the OCP are also listed. The results show that the NN compensator attenuated the error of the nonlinear model by more than 50\%. The acceleration RMSE was largest in the $X$ direction, as this direction contained the largest aerodynamic area. An FFNN with 10 neurons was applied in the experiments, due to its favorable trade-off between predictive accuracy and implementation time.

\begin{table}[hbt!]
	\renewcommand{\arraystretch}{1}
	\caption{Comparison of model performances in terms of OCP solving time and the RMSEs of acceleration predictions.}
	\label{train_rmse}
 \begin{center}
 	\begin{tabular}{>{\centering\arraybackslash}m{5em}
  >{\centering\arraybackslash}m{7em}
  >{\centering\arraybackslash}m{3em}
  >{\centering\arraybackslash}m{3em}
  >{\centering\arraybackslash}m{3em}}
		\hline 
            Architecture & Solving Time (\si{ms}) & X (\si{m/s^2}) & Y (\si{m/s^2}) & Z (\si{m/s^2})\\
		\hline 
		Nonlinear & 5.5 & 0.640 & 0.186 & 0.544\\
            5 neurons & 8 & 0.574 & 0.105 & 0.286\\
            10 neurons & 9.5 & 0.363 & 0.105 & 0.281\\
            100 neurons & 20 & 0.279 & 0.097 & 0.231\\
            500 neurons & 42 & 0.245 & 0.089 & 0.208\\
            \hline
	\end{tabular}
  \end{center}
\end{table}

\begin{figure}[thpb]
	\centering
	\includegraphics[width=0.48\textwidth]{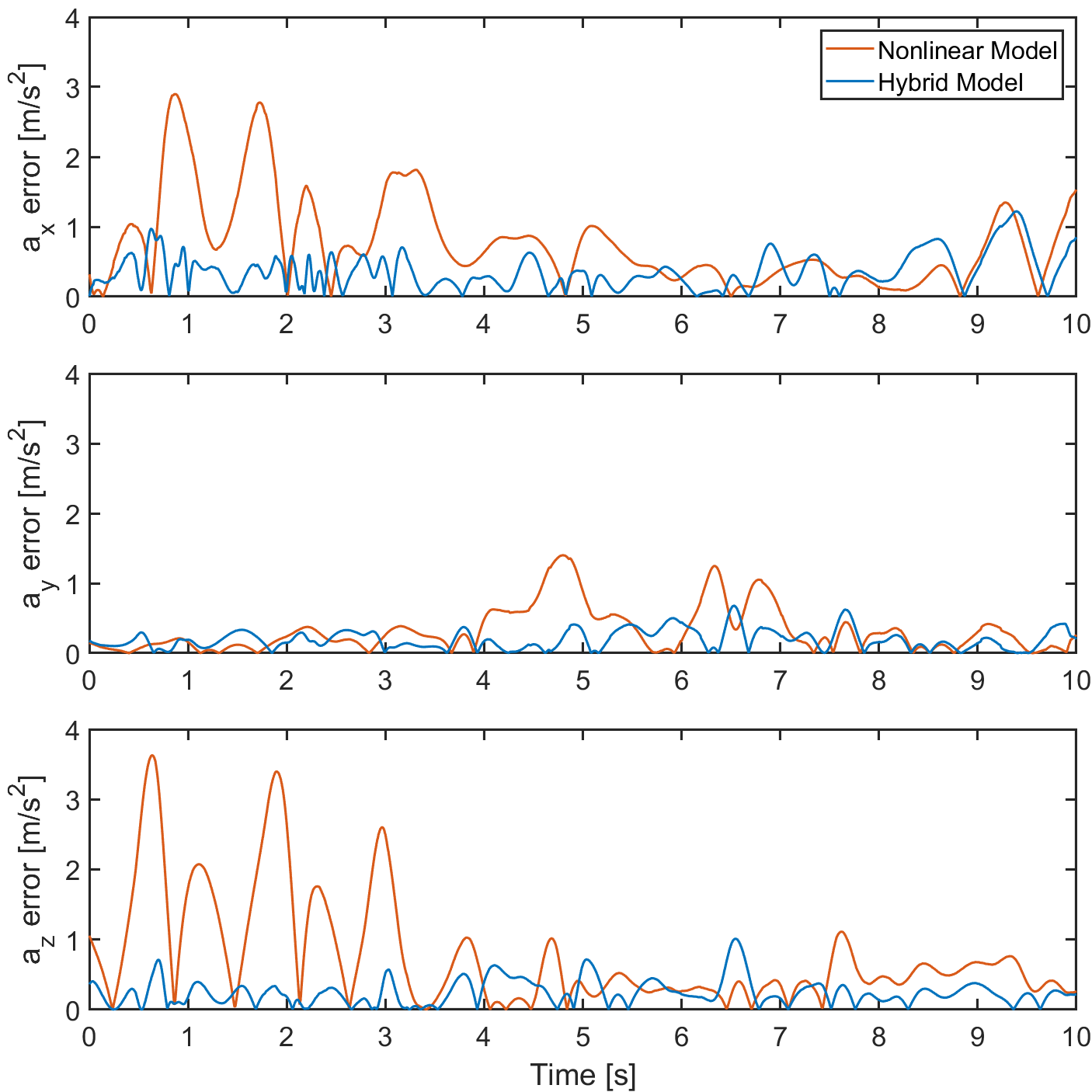}
	\caption{Errors in the acceleration predictions generated by the nonlinear and hybrid models from the testing dataset.}
	\label{model_performance}
\end{figure}

The errors in the predictions generated by the nonlinear and hybrid models from the testing dataset are shown in Fig. \ref{model_performance}. The results show that the error in the predictions generated by the nonlinear model was as high as 3 \si{m/s^2}. Such an error in acceleration prediction would be amplified through integration during position prediction and result in large position-tracking errors. In contrast, the hybrid model obtained a much better predictive accuracy by learning implicit aerodynamics, such as wing drag and propeller--wing interactions, which are especially significant during large maneuvers. Thus, the errors in the predictions generated by the hybrid model were less than 1 \si{m/s^2}. A proportion of this was due to noise, which might have been sensor measurement noise or disturbances from random turbulent airflow. By providing enough training samples, the NN did not overfit the noisy training data and learnt only the underlying aerodynamic effects.

\subsection{Experimental Results}

In flight testing, MPC was implemented on the onboard computer mentioned in Section \ref{experiment_setup} with ACADOS toolbox to generate a fast C-code solver. The NN model function was generated by MATLAB $genFunction$ syntax and used for HMPC. The MPC ran with a 100-Hz update frequency and $20\times0.05${\si{s}} horizon, as shown in Table \ref{mpc_parameters}, which yielded an OCP solving time of less than 15 {\si{ms}} with an approximately 20\% CPU load.

The diagonal weighting matrices $\mathbf{Q}$ and $\mathbf{Q}_N$ correspond to $[\mathbf{x}\ \mathbf{u}]$ and $\mathbf{x}$, respectively. A higher weighting was applied to position than to velocity, to favor position tracking accuracy over transient responsiveness. A relatively high penalty was applied to control inputs to prevent the excitation of inner-loop controllers with low position-tracking error. 

\begin{table}[hbt!]
	\renewcommand{\arraystretch}{1}
	\caption{MPC parameters and weights in the cost function}
	\label{mpc_parameters}
 \begin{center}
	\begin{tabular}{cc}
		\hline 
		MPC Parameter & Value \\
		\hline 
		Prediction horizon & 20\\
            Sample time (\si{s}) & 0.05\\
            $\mathbf{Q}$ & diag [12 12 12 3 3 3 1 1 400 30 30]\\
            $\mathbf{Q}_N$ & diag [12 12 12 3 3 3 1 1]\\
		\hline 
	\end{tabular} 
  \end{center}
\end{table}

\begin{figure}[thpb]
	\centering
	\includegraphics[width=0.48\textwidth]{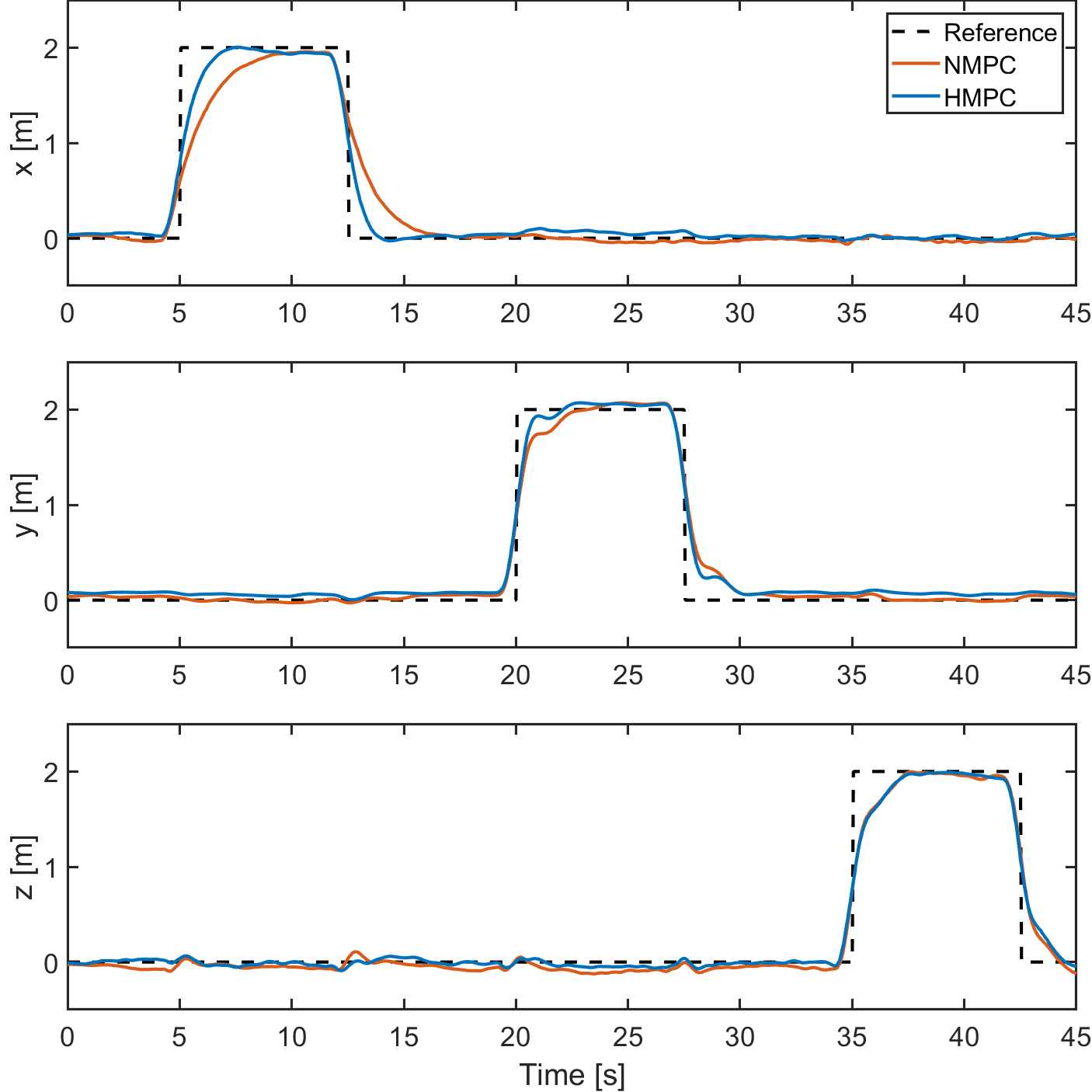}
	\caption{Comparison of step responses of NMPC with those of HMPC}
	\label{step_response}
\end{figure}

To investigate how MPC benefited from the NN-augmented model, we compared the tracking performance of MPC for different trajectories. Both NMPC and HMPC use the same set of parameters. The step responses of NMPC and HMPC are shown in Fig. \ref{step_response} and show that the hybrid model exhibited better tracking performance in the $X$ direction, as the residual dynamics were greater in this direction than in the $Y$ and $Z$ directions. With the application of the hybrid model, the rise time decreased from 3.7 \si{s} to 2.0 \si{s} and the positional RMSE decreased from 0.106 \si{m} to 0.071 \si{m}.

\begin{figure}[thpb]
	\centering
	\includegraphics[width=0.48\textwidth]{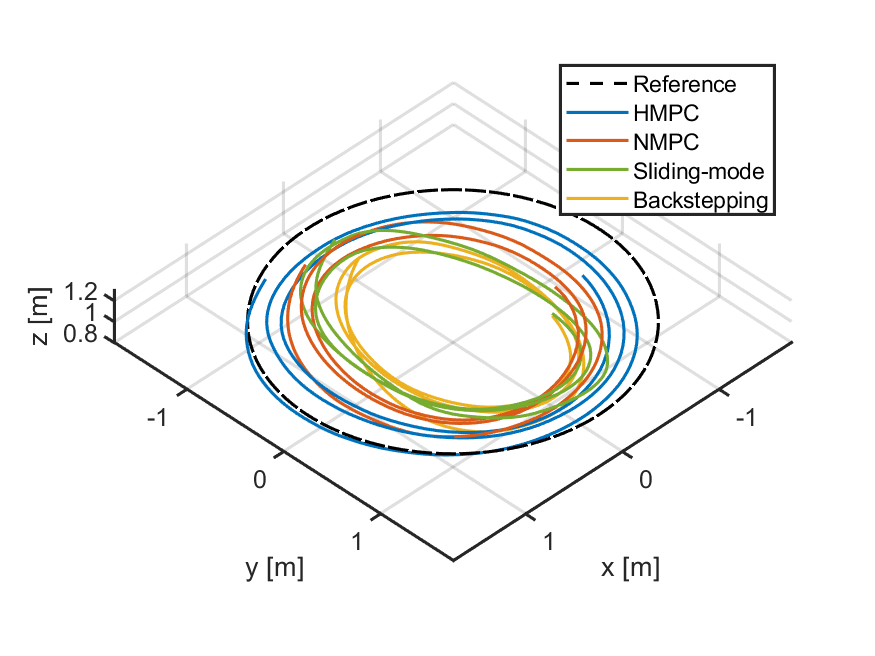}
	\caption{Circular path tracking results of HMPC, NMPC, backstepping, and sliding-mode controllers.}
	\label{circle_trajectory}
\end{figure}

\begin{figure}[thpb]
	\centering
	\includegraphics[width=0.48\textwidth]{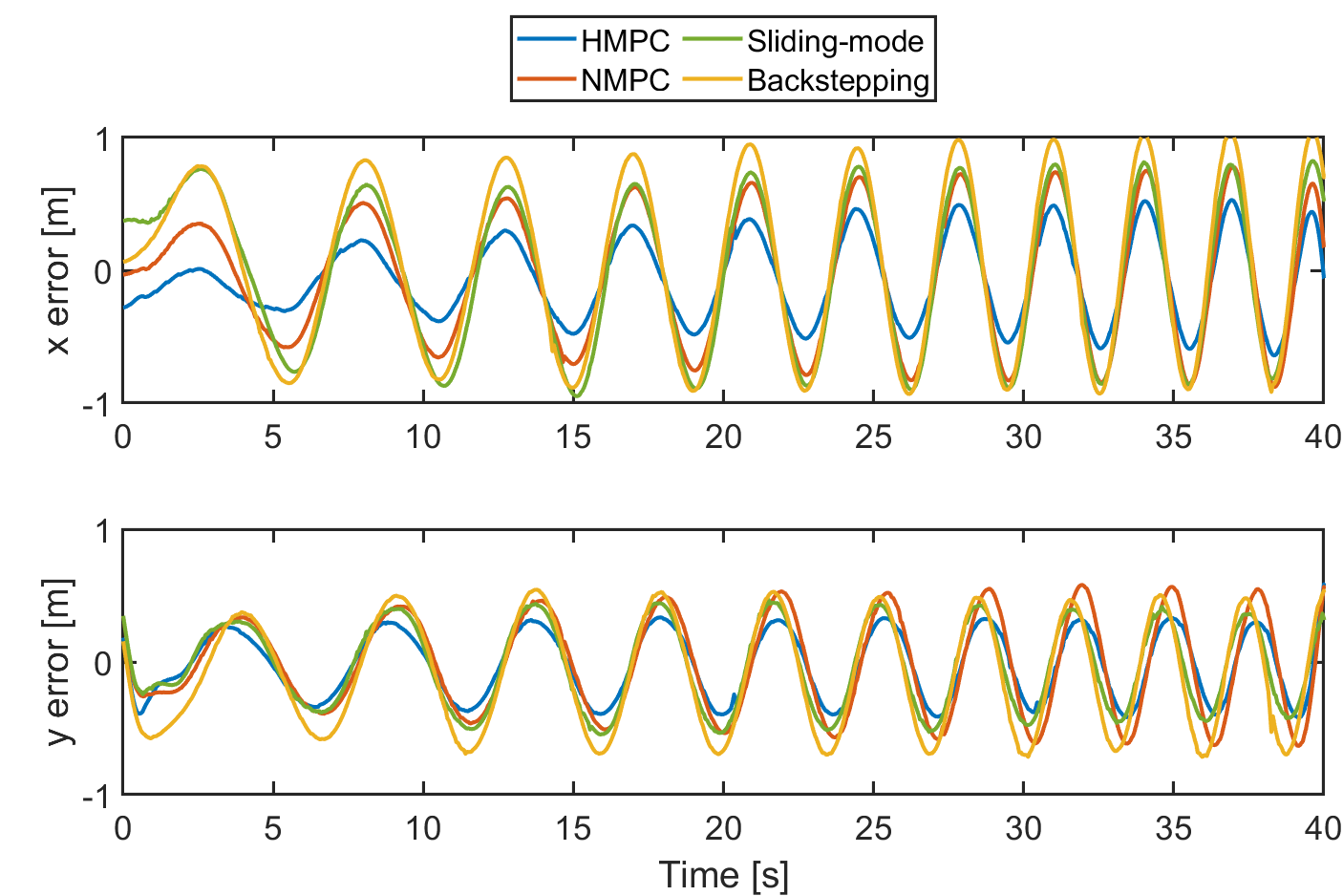}
	\caption{Circular path tracking errors of HMPC, NMPC, backstepping, and sliding-mode controllers at velocity up to $3 m/s$.}
	\label{circle_error}
\end{figure}

The maximum velocity achieved in the step response was approximately $1 m/s$ and the residual dynamics remained nominal in this scenario. Therefore, a high-speed circular trajectory not included in the training dataset was designed to investigate the tracking performance in a more challenging scenario. The trajectory was a circle in the $XY$ plane with a diameter of $3m$, and the reference speed was increased from $1.5m/s$ to $3m/s$ in $40$ seconds. Figure \ref{circle_trajectory} depicts the tracked trajectories of HMPC, NMPC, backstepping, and sliding-mode controllers, showing that the hybrid model improved the tracking accuracy in both $X$ and $Y$ directions. That is, NMPC failed to accurately predict the residual aerodynamic effects and thus displayed an elliptical trajectory, while HMPC generated a near-circular trajectory. The position-tracking error of the circular trajectory is shown in Fig. \ref{circle_error}. indicating that the benchmark nonlinear control methods result in larger tracking errors compared to HMPC, especially in $X$ direction where the unmodeled dynamics has the largest influence on the system. Apart from the circular trajectory, the controller performance is also evaluated by an unseen figure-eight (lemniscate) trajectory described by

\begin{linenomath}\begin{equation}
	\label{lemniscate_track}
	\left\{
	\begin{aligned}
		\ x &=2cos(1.5t)\\
		\ y &=2cos(1.5t)sin(1.5t)\\
	\end{aligned}
	\right..
\end{equation}\end{linenomath}

\begin{figure}[thpb]
	\centering
	\includegraphics[width=0.48\textwidth]{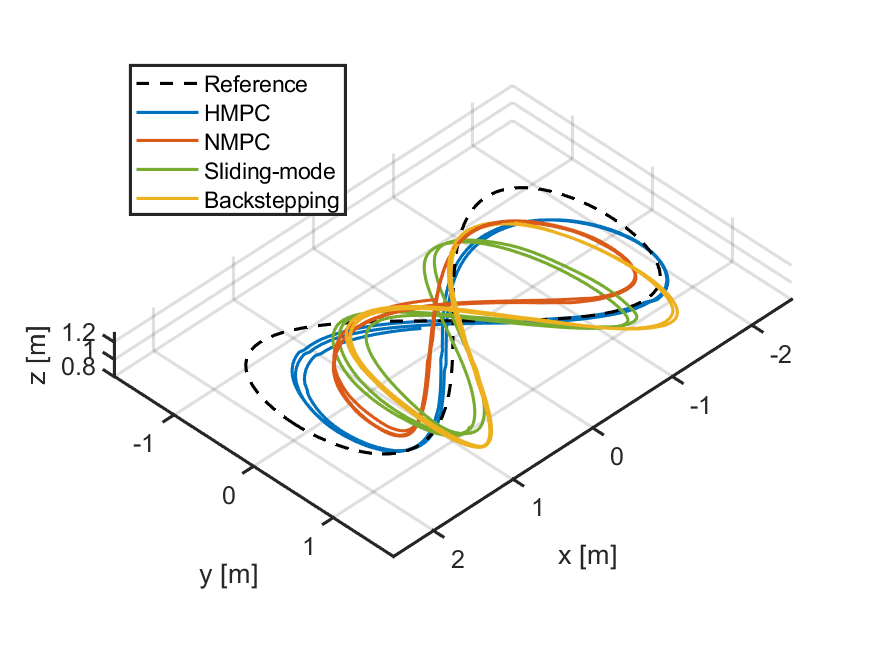}
	\caption{Figure-eight path tracking results of HMPC, NMPC, backstepping, and sliding-mode controllers.}
	\label{lemniscate_trajectory}
\end{figure}

\begin{figure}[thpb]
	\centering
	\includegraphics[width=0.48\textwidth]{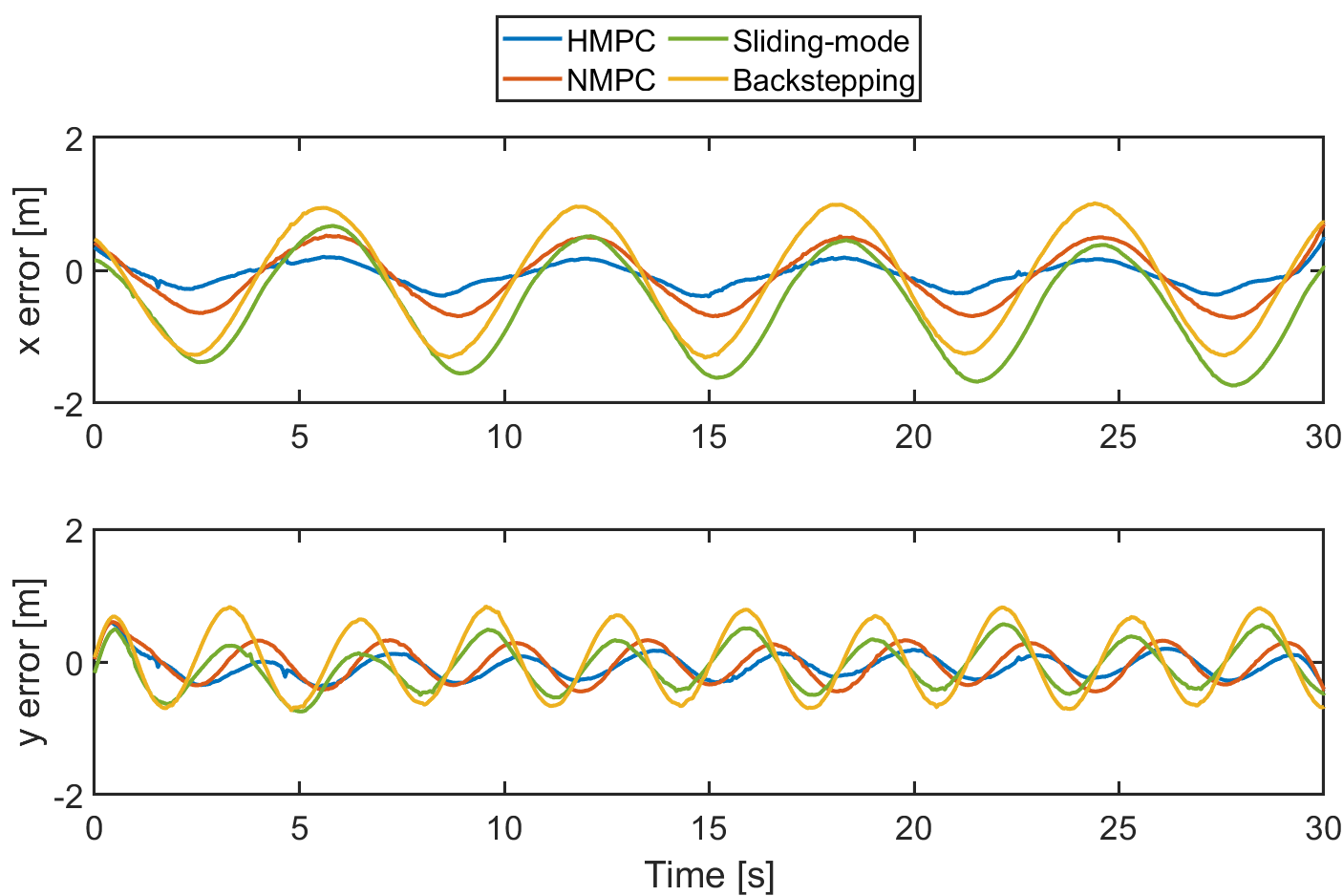}
	\caption{Figure-eight path tracking error of HMPC, NMPC, backstepping, and sliding-mode controllers.}
	\label{lemniscate_error}
\end{figure}

The tracking results are shown in Fig.\ref{lemniscate_trajectory} and Fig.\ref{lemniscate_error}. The nonlinear benchmark controllers can barely follow the highly dynamic reference trajectory resulting in a distorted trajectory shape, while the HMPC is able to follow most parts of the trajectory. Besides, the tracking results of both sliding-mode and backstepping control contain large overshoots. The RMSE of position tracking is summarized in Table \ref{tracking_rmse}, indicating 37\% and 57\% reductions in tracking error of HMPC compared to NMPC on circular and figure-eight trajectories, respectively. Note that since the PID controllers result in large tracking errors that is not with the same level of other nonlinear control methods, the performance is not plotted in the previous results. In general, the HMPC has the lowest tracking errors, especially in $X$ direction where the dynamics is the most complicated.

An HMPC with a reduced training dataset is introduced to evaluate the generalization performance of the proposed HMPC. The reduced training dataset only covers translational velocity up to $2m/s$. The original dataset contains around $150k$ sampling points while the reduced one only contains $100k$. The tracking RMSE is shown in Table \ref{tracking_rmse} by HMPC*. The results show that HMPC performance degrades with the reduced training dataset but still outperforms NMPC, which indicates that the proposed approach is capable of generalize the dynamics beyond the data it was trained on. The tracked trajectory of HMPC* is not plotted in previous figures since it is very similar to that of HMPC.

\begin{table}[hbt!]
	\renewcommand{\arraystretch}{1}
	\caption{Comparison of the position-tracking RMSE on circular and figure-eight trajectory.}
	\label{tracking_rmse}
 \begin{center}
 	\begin{tabular}{>{\centering\arraybackslash}m{6em}
  >{\centering\arraybackslash}m{2.8em}
  >{\centering\arraybackslash}m{2.8em}
  >{\centering\arraybackslash}m{2.8em}
  >{\centering\arraybackslash}m{2.8em}}
		\hline 
		\multirow{2}{6em}{Model type}	& \multicolumn{2}{c}{Circular Trajectory} & \multicolumn{2}{c}{Figure-eight Trajectory}\\
            & x (\si{m}) & y (\si{m})& x (\si{m}) & y (\si{m})\\
		\hline
            PID & 0.958 & 1.103 & 1.247 & 0.853\\
            Backstepping  & 0.573 & 0.366 & 0.696 & 0.446\\
		Sliding-mode  & 0.513 & 0.277 & 0.729 & 0.284\\
            NMPC & 0.413 & 0.316 & 0.372 & 0.232\\
            HMPC* & 0.268 & 0.229 & 0.203 & 0.197\\
            HMPC  & \textbf{0.259} & \textbf{0.225 }& \textbf{0.159 }&\textbf{0.158}\\
            \hline
	\end{tabular}
  \end{center}
\end{table}

\begin{figure}[thpb]
	\centering
  	\includegraphics[width=0.45\textwidth]{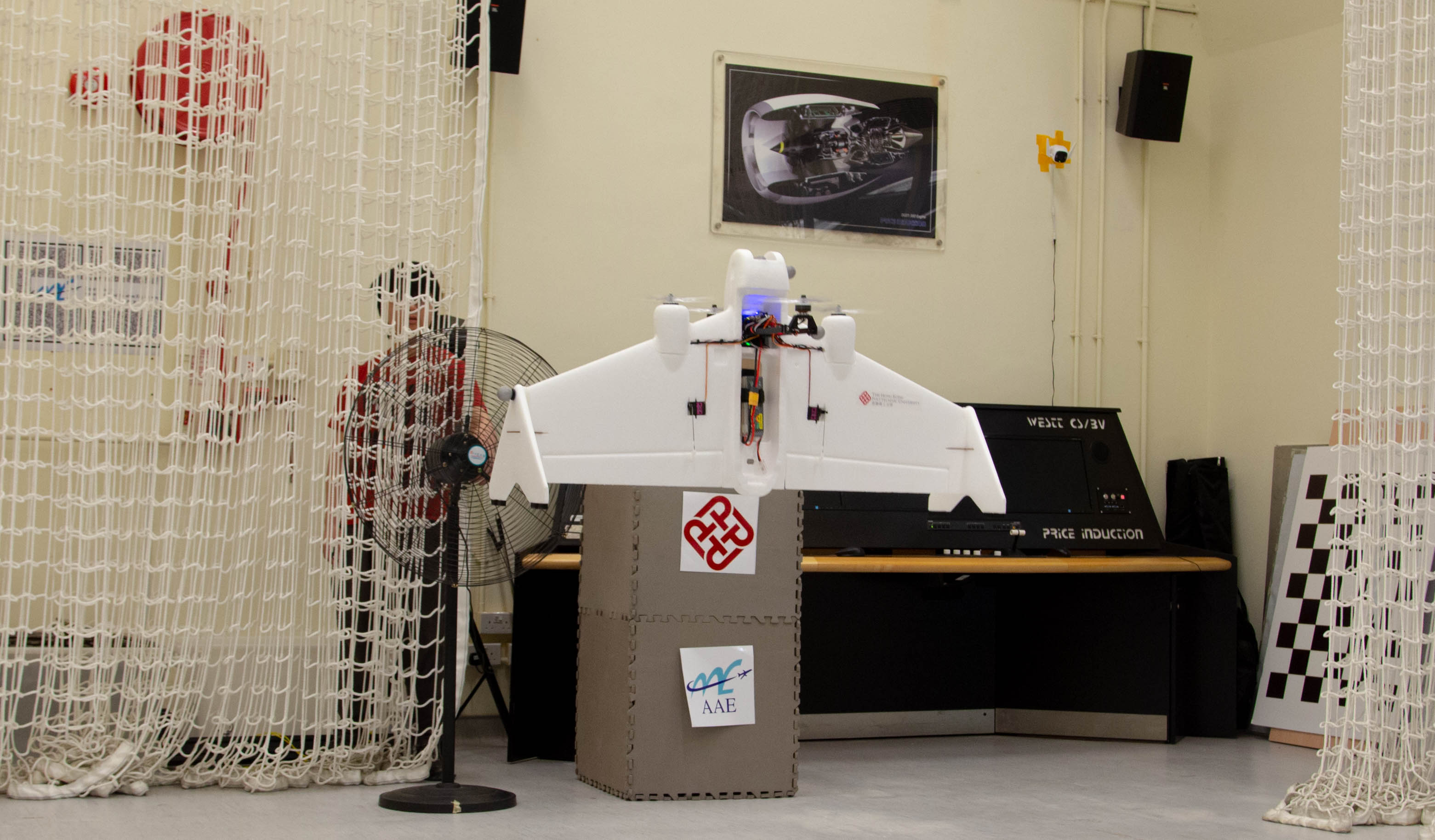}
	\includegraphics[width=0.48\textwidth]{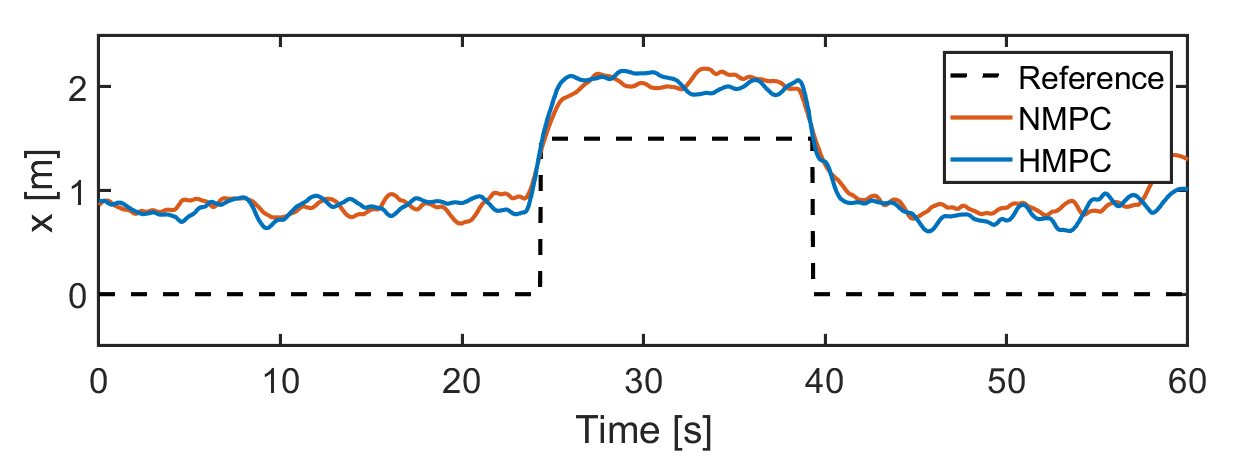}
	\caption{The upper figure shows the setup of the experiment in which a wind disturbance was applied onto the wing of the quadrotor tail-sitter UAV, while the lower figure shows the X-direction step response for a wind of up to $3 m/s$.}
	\label{wind_disturbance}
\end{figure}

As an NN is a black-box model, we also tested the robustness of HMPC by introducing a wind disturbance of approximately 3 \si{m/s} in the $X$ direction, which was the most challenging direction. The experimental setup and the step response under this wind disturbance are shown in Fig. \ref{wind_disturbance}. The NN does not consider any external disturbances, and thus the tracking errors of NMPC and HMPC were similar.

\section{Conclusion}

This study developed an NN-based hybrid aerodynamic model of a quadrotor tail-sitter UAV that exhibits good accuracy and a low computational cost. HMPC for position tracking in hovering flight was implemented based on this model and was validated in real-world flight experiments. HMPC was strongly generalizable to unseen trajectories and achieved a 57\% reduction in the tracking error relative to NMPC. The robustness of HMPC was confirmed in the presence of a wind disturbance of up to $3 m/s$. This approach simplifies the modeling and identification process of a quadrotor tail-sitter UAV without requiring wind tunnel experiments or CFD calculations. Future work could extend this approach to encompass transition and cruise stages of UAV operation. Additionally, there is potential for exploring the feasibility of training the model online, allowing it to adapt dynamically to changing conditions, such as a mass change or variable wind disturbances.

\bibliographystyle{IEEEtaes}
\bibliography{references}

\begin{IEEEbiography}[{\includegraphics[width=1in,height=1.25in,clip,keepaspectratio]{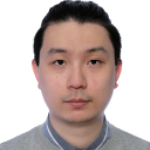}}]{Bailun JIANG} received his BEng and MSc in Mechanical Engineering from The Hong Kong Polytechnic University in 2017 and 2020, Hong Kong, China.

He is currently a member in the AIRo-Lab, Department of Aeronautical and Aviation Engineering, The Hong Kong Polytechnic University. His research interests include model predictive control, deep learning, and path planning of UAVs.
\end{IEEEbiography}

\begin{IEEEbiography}[{\includegraphics[width=1in,height=1.25in,clip,keepaspectratio]{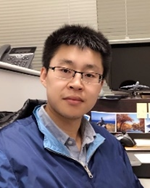}}]{Boyang LI (M'16)} received his B.Eng. and M.Eng. degrees in Aeronautical Engineering from Northwestern Polytechnical University, Xi’an, China, in 2012 and 2015. He then received PhD degree in Mechanical Engineering from The Hong Kong Polytechnic University in 2019. 

He has conducted postdoctoral research at the Air Traffic Management Research Institute, Nanyang Technological University, Singapore and the School of Engineering, the University of Edinburgh, UK. In 2020, he established the Autonomous Aerial Systems Lab at The Hong Kong Polytechnic University. He joined the University of Newcastle as a lecturer in Aerospace Systems Engineering in 2023. His research interests include model predictive control, path/trajectory optimization, and field experiments of unmanned aircraft systems (UAS) and other mobile robots.
\end{IEEEbiography}

\begin{IEEEbiography}[{\includegraphics[width=1in,height=1.25in,clip,keepaspectratio]{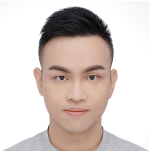}}]{Ching-Wei Chang} received his Bachelor of Engineering degree from the Department of Mechanical Engineering at Yuan Ze University, Taiwan, in 2015. He then received Ph.D. degree in Mechanical Engineering from The Hong Kong Polytechnic University in 2022. His research interests include VTOL UAVs, UAV inspection applications, and autonomous landing systems for UAVs.
\end{IEEEbiography}

\begin{IEEEbiography}[{\includegraphics[width=1in,height=1.25in,clip,keepaspectratio]{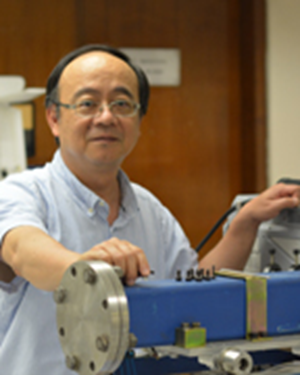}}]{Chih-Yung WEN} received his B.S. degree from the Department of Mechanical Engineering at National Taiwan University in 1986 and his M.S. and Ph.D. degrees from the Department of Aeronautics at the California Institute of Technology (Caltech) in 1989 and 1994, respectively.

He joined the Department of Mechanical Engineering, The Hong Kong Polytechnic University in 2012, as a professor. In 2021, he became the chair professor and head of the Department of Aeronautical and Aviation Engineering in The Hong Kong Polytechnic University. His current research interests include modeling and control of tail-sitter UAVs, visual-inertial odometry systems for UAVs and AI object detection by UAVs.
\end{IEEEbiography}
\vfill

\end{document}